\documentclass{article}

\usepackage{arxiv}
\usepackage{bm}
\usepackage[utf8]{inputenc} % allow utf-8 input
\usepackage[T1]{fontenc}    % use 8-bit T1 fonts
\usepackage{hyperref}       % hyperlinks
\usepackage{url}            % simple URL typesetting
\usepackage{booktabs}       % professional-quality tables
\usepackage{amsmath,amssymb,amsfonts,amsthm,mathtools,array}
\usepackage{nicefrac}       % compact symbols for 1/2, etc.
\usepackage{microtype}      % microtypography
\usepackage{cleveref}       % smart cross-referencing
\usepackage{autonum}
\usepackage{graphicx}
\usepackage{color}
\usepackage{natbib}
\usepackage{doi}
\usepackage{breakurl}

\title{Mathematical Definition and Systematization of Puzzle Rules}

\newif\ifuniqueAffiliation
\uniqueAffiliationtrue

\ifuniqueAffiliation % Standard variant of author block
\author{ {\hspace{1mm}Itsuki Maeda} \\
	Department of Micro Engineering,\\
	Kyoto University\\
	Kyoto city, Kyoto 615-8540 JAPAN \\
	\texttt{maeda.itsuki.b27@kyoto-u.jp} \\
	\And
	\href{https://researchmap.jp/yasuhiroinoue}{\hspace{1mm}Yasuhiro Inoue} \\
	Department of Micro Engineering,\\
	Kyoto University\\
	Kyoto city, Kyoto 615-8540 JAPAN \\
	\texttt{inoue.yasuhiro.4n@kyoto-u.ac.jp} \\
}
\else
\usepackage{authblk}

\newbox{\orcid}\sbox{\orcid}{}
\author[1]{%
	\href{https://orcid.org/0000-0000-0000-0000}{\usebox{\orcid}\hspace{1mm}David S.~Hippocampus\thanks{\texttt{hippo@cs.cranberry-lemon.edu}}}%
}
\author[1,2]{%
	\href{https://orcid.org/0000-0000-0000-0000}{\usebox{\orcid}\hspace{1mm}Elias D.~Striatum\thanks{\texttt{stariate@ee.mount-sheikh.edu}}}%
}
\affil[1]{Department of Computer Science, Cranberry-Lemon University, Pittsburgh, PA 15213}
\affil[2]{Department of Electrical Engineering, Mount-Sheikh University, Santa Narimana, Levand}
\fi

\renewcommand{\shorttitle}{Mathematical Definition and Systematization of Puzzle Rules}

\hypersetup{
pdftitle={Mathematical Definition and Systematization of Puzzle Rules},
pdfauthor={Itsuki Maeda, Yasuhiro Inoue},
}

\newtheorem{definition}{Definition}[section]
\newtheorem{example}{Example}[section]
\newtheorem{remark}{Remark}[section]
\definecolor{myred}{RGB}{255,50,50}         % revision version

\def\emptyline{\vskip.5\baselineskip}
\def\dcomma{\,\,}

\crefname{section}{Section}{Sections}
\Crefname{section}{Section}{Sections}

\crefname{chapter}{Chapter}{Chapters}
\Crefname{chapter}{Chapter}{Chapters}

\crefname{figure}{Figure}{Figures}
\Crefname{figure}{Figure}{Figures}
\crefname{table}{Table}{Tables}
\Crefname{table}{Table}{Tables}

\crefname{equation}{Equation}{Equations}
\Crefname{equation}{Equation}{Equations}
\crefname{theorem}{Theorem}{Theorems}
\Crefname{theorem}{Theorem}{Theorems}

\begin{document}
\maketitle

\begin{abstract}
	While logic puzzles have engaged individuals through problem-solving and critical thinking, the creation of new puzzle rules has largely relied on ad-hoc processes. Pencil puzzles, such as Slitherlink and Sudoku, represent a prominent subset of these games, celebrated for their intellectual challenges rooted in combinatorial logic and spatial reasoning. Despite extensive research into solving techniques and automated problem generation, a unified framework for systematic and scalable rule design has been lacking. Here, we introduce a mathematical framework for defining and systematizing pencil puzzle rules. This framework formalizes grid elements, their positional relationships, and iterative composition operations, allowing for the incremental construction of structures that form the basis of puzzle rules. Furthermore, we establish a formal method to describe constraints and domains for each structure, ensuring solvability and coherence. Applying this framework, we successfully formalized the rules of well-known Nikoli puzzles, including Slitherlink and Sudoku, demonstrating the formal representation of a significant portion (approximately one-fourth) of existing puzzles. These results validate the potential of the framework to systematize and innovate puzzle rule design, establishing a pathway to automated rule generation. By providing a mathematical foundation for puzzle rule creation, this framework opens avenues for computers, potentially enhanced by AI, to design novel puzzle rules tailored to player preferences, expanding the scope of puzzle diversity. Beyond its direct application to pencil puzzles, this work illustrates how mathematical frameworks can bridge recreational mathematics and algorithmic design, offering tools for broader exploration in logic-based systems, with potential applications in educational game design, personalized learning, and computational creativity.
\end{abstract}

% keywords can be removed

\section{Introduction}\label{section:Introduction}

Puzzles have long been a source of intellectual engagement and entertainment, captivating people worldwide through their ability to foster problem-solving and critical thinking skills. Among the diverse genres of puzzles, pencil puzzles, such as Slitherlink and Sudoku, are particularly celebrated for their logical depth and reliance on combinatorial reasoning and spatial intuition. These puzzles, as deterministic games with perfect information, have been extensively studied in terms of their computational complexity, solving algorithms, and automated problem generation. However, despite these advances, the creation of new puzzle rules has remained a largely ad-hoc process, lacking a unified and systematic approach.

The need for automating puzzle rule creation arises from practical challenges in maintaining engagement and diversity in puzzle-based games. \citeauthor{deKegel2020} observed that:
“Players tend to feel a sense of monotony from repetitive gameplay. However, this can be mitigated by introducing puzzles.” (\citealt{deKegel2020}).

Offering a variety of puzzles, particularly those with novel rules, is widely regarded as an effective way to sustain player engagement. While specific studies directly supporting this notion are limited, \citeauthor{Tang2024} demonstrated that curiosity-driven design, including the introduction of uncertainty and novel challenges, plays a critical role in maintaining player interest across diverse gaming contexts (\citealt{Tang2024}). Their findings emphasize the importance of catering to multiple dimensions of curiosity, such as epistemic and perceptual curiosity, to enhance engagement and foster loyalty in players. However, designing a wide variety of puzzle rules often involves significant production costs, making it challenging to achieve diversity efficiently. To address these issues, there is a clear demand for frameworks that enable the automatic generation of puzzle rules, which could facilitate the creation of diverse and innovative challenges.

During our literature review, we found no studies that generated puzzle rules entirely from scratch. A closely related study explored the generation of two-player finite deterministic perfect information games by combining existing games (\citealt{Browne2010}). However, this approach focused on combining pre-existing rules rather than creating new ones, and it relied on engineering methods instead of a mathematical framework. Other researches (\citealt{Mantere2007}; \citealt{Yoshinaka2012}) have concentrated on generating specific puzzle instances  that adhere to established rules, rather than defining new rules themselves.  \citeauthor{Herting2004} proposed a mathematical formulation for the board and conditions of the Slitherlink puzzle, employing a rule-based approach for efficient problem-solving, but this work did not address the generation of new rules (\citealt{Herting2004}). While existing studies have addressed puzzle solving and problem generation, the systematic creation of novel pencil puzzle rules from a mathematical framework has not been extensively investigated, to the best of our knowledge.

In this study, we address the lack of systematic methods for creating pencil puzzle rules by introducing a comprehensive mathematical framework. Our framework formalizes grid elements (e.g., points, edges, cells) and their relationships (e.g., adjacency) and iteratively composes them into structures that form the foundation of puzzle rules. Additionally, it provides a formal method for defining constraints and solution domains, ensuring solvability and coherence of the puzzles. This systematic approach enables the mathematical description and computational implementation of puzzle rules, offering a unified foundation for rule creation across diverse puzzle types.

This framework has broad implications beyond pencil puzzle design. Automated rule generation could enhance commercial games, personalize learning experiences, and contribute to fields like network security, where logic-based systems play a crucial role (\citealt{Liang2013}). Additionally, the framework opens avenues for future research into AI-driven rule generation and personalized puzzle design tailored to player preferences. By bridging recreational mathematics and algorithmic design, this work establishes a foundation for innovations in both entertainment and academic contexts.

In this paper, we begin by defining key terms and concepts related to pencil puzzles and their rules in In \cref{section:MathematicalDefinition}. Using these definitions, we demonstrate the framework's validity in \cref{section:verification} by applying it to well-known puzzles and verifying its effectiveness through computational implementation. In \cref{section:Conclusion}, we summarize this study by revisiting the key results and discussing their implications, including the potential for systematic puzzle rule design.

\section{Mathematical Definitions of Basic Concepts Related to Pencil Puzzles}\label{section:MathematicalDefinition}
First, consider a two-dimensional grid of size $m\times n$ ($m,n\in\mathbb{N}$). In this case, we can assign coordinates $(i,j)$ to each grid point from the top-left corner ($1\le i \le m+1,, 1\le j \le n+1$). Similarly, we can assign coordinates $(i', j')$ to each cell from the top-left ($1\le i' \le m,, 1\le j' \le n$). Then, we can give annotations with variables $i, j, i', j'$ to grid points, cells, horizontal edges connecting grid points, vertical edges connecting grid points, horizontal edges connecting cells, and vertical edges connecting cells in the two-dimensional grid.

We define these annotations as \textbf{grid point} $p(i,j)$, \textbf{cell} $c(i',j')$, \textbf{grid point horizontal edge} $h_p(i,j)$, \textbf{grid point vertical edge} $v_p(i,j)$, \textbf{cell horizontal edge} $h_c(i',j')$, and \textbf{cell vertical edge} $v_c(i',j')$, and collectively refer to these as \textbf{elements} In this study. Furthermore, we will refer to sequences containing these as grid point sequence, cell sequence, grid point horizontal edge sequence, grid point vertical edge sequence, cell horizontal edge sequence, and cell vertical edge sequence.

\emptyline
\begin{definition}[Grid point sequence $\mathbf{P}$, Cell sequence $\mathbf{C}$, Grid point horizontal edge sequence $\mathbf{H_p}$, Grid point vertical edge sequence $\mathbf{V_p}$, Cell sequence $\mathbf{C}$, Cell horizontal edge sequence $\mathbf{H_c}$, Cell vertical edge sequence $\mathbf{V_c}$, Grid point edge sequence $\mathbf{E_p}$, Cell edge sequence $\mathbf{E_c}$, Element sequence $\mathbb{E}$]\label{definition:Elements}
	Given a plane grid of size $m\times n$, we define the \textbf{grid point sequence} $\mathbf{P}$ as
	\begin{equation}
		\mathbf{P}\coloneqq\bigl(\,p(1,1), \dots,p(m+1,n+1) \,\bigr)
	\end{equation}
	Similarly, we define \textbf{grid point horizontal edge sequence} $\mathbf{H_p}$, \textbf{grid point vertical edge sequence} $\mathbf{V_p}$, \textbf{cell sequence} $\mathbf{C}$, \textbf{cell horizontal edge sequence} $\mathbf{H_c}$, \textbf{cell vertical edge sequence} $\mathbf{V_c}$, \textbf{grid point edge sequence} $\mathbf{E_p}$, \textbf{cell edge sequence} $\mathbf{E_c}$ as
	\begin{alignat}{2}
		 & \mathbf{C  } & \coloneqq \dcomma & \bigl(\,c(1,1), \dots,c(m,n)\,\bigr)        \\
		 & \mathbf{H_p} & \coloneqq \dcomma & \bigl(\,h_p(1,1), \dots,h_p(m+1,n) \,\bigr) \\
		 & \mathbf{V_p} & \coloneqq \dcomma & \bigl(\,v_p(1,1), \dots,v_p(m,n+1) \,\bigr) \\
		 & \mathbf{H_c} & \coloneqq \dcomma & \bigl(\,h_c(1,1), \dots,h_c(m,n-1) \,\bigr) \\
		 & \mathbf{V_c} & \coloneqq \dcomma & \bigl(\,v_c(1,1), \dots,v_c(m-1,n) \,\bigr) \\
		 & \mathbf{E_p} & \coloneqq \dcomma & \mathbf{H_p} \cup \mathbf{V_p}              \\
		 & \mathbf{E_c} & \coloneqq \dcomma & \mathbf{H_c} \cup \mathbf{V_c}
	\end{alignat}
	Also, we define the \textbf{element sequence} $\mathbb{E}$ composed of these sequences as
	\begin{equation}
		\mathbb{E}\coloneqq(\,\mathbf{P} ,\mathbf{E_p}, \mathbf{C}, \mathbf{E_c}\,)
	\end{equation}
\end{definition}
\emptyline
Next, we introduce relationships when two arbitrary elements are taken from the two-dimensional grid. This defines positional relationships for annotations from board information to variables as defined earlier. Here, positional relationships are binary predicates in the context of mathematical logic that take any element included in $\mathbb{E}$ as arguments. We define and introduce these relationships as horizontal adjacency, vertical adjacency, diagonal adjacency, and coincidence below.
\emptyline
\begin{definition}[Horizontal adjacency $\mathrm{H}(x,y)$, Vertical adjacency $\mathrm{V}(x,y)$, Diagonal adjacency $\mathrm{D}(x,y)$, Coincidence $\mathrm{M}(x,y)$]\label{definition:PositionalRelationship}
	We define
	\textbf{horizontal adjacency} $\mathrm{H}(x,y)$ as

	\begin{equation}
		\mathrm{H}(x, y) \coloneqq
		\begin{cases}
			\bigl[\,\,i = i' \land |j - j'| = 1 \quad \mathrm{if} \quad x = p(i, j) \land y = p(i', j')            \,\bigr]     \\
			\lor                                                                                                                \\
			\bigl[\,\,i = i' \land |j - j'| = 1 \quad \mathrm{if} \quad x = c(i, j) \land y = c(i', j')            \,\bigr]     \\
			\lor                                                                                                                \\
			\bigl[\,\,i = i' \land |j - j'| = 1 \quad \mathrm{if} \quad x = h_p(i, j) \land y = h_p(i', j')        \,\bigr]     \\
			\lor                                                                                                                \\
			\bigl[\,\,i = i' \land |j - j'| = 1 \quad \mathrm{if} \quad x = h_c(i, j) \land y = h_c(i', j')        \,\bigr]     \\
			\lor
			\\
			\bigl[\,\,i = i' \land  j-j'\in \{\,\,0,1\,\,\} \quad \mathrm{if} \quad x = p(i, j) \land y = h_p(i', j')  \,\bigr] \\
			\lor                                                                                                                \\
			\bigl[\,\,i = i' \land  j-j'\in \{\,\,0,-1\,\,\} \quad \mathrm{if} \quad x = h_p(i, j) \land y = p(i', j') \,\bigr] \\
			\lor                                                                                                                \\
			\bigl[\,\,i = i' \land  j-j'\in \{\,\,0,1\,\,\} \quad \mathrm{if} \quad x = c(i, j) \land y = h_c(i', j')  \,\bigr] \\
			\lor                                                                                                                \\
			\bigl[\,\,i = i' \land  j-j'\in \{\,\,0,-1\,\,\} \quad \mathrm{if} \quad x = h_c(i, j) \land y = c(i', j')\,\bigr].
		\end{cases}
	\end{equation}

	Similarly, we define \textbf{vertical adjacency} $\mathrm{V}(x,y)$, \textbf{diagonal adjacency} $\mathrm{D}(x,y)$, and \textbf{coincidence} $\mathrm{M}(x,y)$ as

	\begin{flalign}
		& \mathrm{V}(x, y) \coloneqq
	 \begin{cases}
		 \bigl[\,\,|i-i'| = 1 \land j=j' \quad \mathrm{if} \quad x = p(i, j) \land y = p(i', j')       \,\bigr]             \\
		 \lor                                                                                                               \\
		 \bigl[\,\,|i-i'| = 1 \land j=j' \quad \mathrm{if} \quad x = c(i, j) \land y = c(i', j')       \,\bigr]             \\
		 \lor                                                                                                               \\
		 \bigl[\,\,|i-i'| = 1 \land j=j' \quad \mathrm{if} \quad x = v_p(i, j) \land y = v_p(i', j')   \,\bigr]             \\
		 \lor                                                                                                               \\
		 \bigl[\,\,|i-i'| = 1 \land j=j' \quad \mathrm{if} \quad x = v_c(i, j) \land y = v_c(i', j')   \,\bigr]             \\
		 \lor                                                                                                               \\
		 \bigl[\,\,i-i' \in \{\,\,0,1\,\,\} \land j=j'\quad \mathrm{if} \quad x = p(i, j) \land y = v_p(i', j')    \,\bigr] \\
		 \lor                                                                                                               \\
		 \bigl[\,\,i-i' \in \{\,\,0,-1\,\,\}  \land j=j' \quad \mathrm{if} \quad x = v_p(i, j) \land y = p(i', j') \,\bigr] \\
		 \lor                                                                                                               \\
		 \bigl[\,\,i-i' \in \{\,\,0,1\,\,\}\land j=j'\quad \mathrm{if} \quad x = c(i, j) \land y = v_c(i', j')     \,\bigr] \\
		 \lor                                                                                                               \\
		 \bigl[\,\,i-i' \in \{\,\,0,-1\,\,\}  \land j=j' \quad \mathrm{if} \quad x = v_c(i, j) \land y = c(i', j')\,\bigr]
	 \end{cases}
	 \\
	 \\
		& \mathrm{D}(x, y) \coloneqq
	 \begin{cases}
		 \bigl[\,\,|i-i'| = 1 \land |j-j'|= 1 \quad \mathrm{if} \quad x = p(i, j) \land y = p(i', j')                               \,\bigr]    \\
		 \lor                                                                                                                                   \\
		 \bigl[\,\,|i-i'| = 1 \land |j-j'|= 1 \quad \mathrm{if} \quad x = c(i, j) \land y = c(i', j')                               \,\bigr]    \\
		 \lor                                                                                                                                   \\
		 \bigl[\,\,i-i' \in \{\,\,0,1\,\,\} \land j-j'\in \{\,\,0,-1\,\,\} \quad \mathrm{if} \quad x = h_p(i, j) \land y = v_p(i', j') \,\bigr] \\
		 \lor                                                                                                                                   \\
		 \bigl[\,\,i-i' \in \{\,\,0,-1\,\,\} \land j-j'\in \{\,\,0,1\,\,\} \quad \mathrm{if} \quad x = v_p(i, j) \land y = h_p(i', j') \,\bigr] \\
		 \lor                                                                                                                                   \\
		 \bigl[\,\,i-i' \in \{\,\,0,1\,\,\} \land j-j'\in \{\,\,0,-1\,\,\} \quad \mathrm{if} \quad x = h_c(i, j) \land y = v_c(i', j') \,\bigr] \\
		 \lor                                                                                                                                   \\
		 \bigl[\,\,i-i' \in \{\,\,0,-1\,\,\} \land j-j'\in \{\,\,0,1\,\,\} \quad \mathrm{if} \quad x = v_c(i, j) \land y = h_c(i', j')\,\bigr]
	 \end{cases}
	 \\
	 \\
		& \mathrm{M}(x, y) \coloneqq \bigl[\,\,x=y\,\bigr]
 \end{flalign}

	These relationships are arbitrary binary predicates inspired by existing puzzle rules. Intuitively, as their names suggest, they are binary predicates that become true when free variables are assigned to elements that are adjacent horizontally, vertically, or diagonally on the board, or coincide.
\end{definition}

\emptyline
In pencil puzzles, in addition to the previously defined elements, there are concepts that combine these elements (e.g., "rooms" in Shikaku, "closed curves" in Slitherlink). In this study, we will refer to these as structures. To mathematically handle structures, we define them as an extended concept of elements.

Furthermore, we introduce order relations for various elements and sequences dealt with In this study.
\emptyline
\begin{definition}[Order Relations]\label{definition:OrderRelationship}
	This research only deals with elements and sequences composed of them. Therefore,
	\emptyline
	\begin{itemize}
		\item The order relation between elements of the same type is determined by coordinates, and for different types, we define $p< c<h_p<v_p<h_c<v_c$.
		\item The order relation between sequences is such that deeper partial sequences are considered larger.
	\end{itemize}
	\emptyline
	We define such order relations. Note that elements can be defined as sequences of depth 0 when the depth of non-nested sequences is considered 1.
\end{definition}
\emptyline
\begin{example}[Order Relations]
	The following group of propositions
	\emptyline
	\begin{itemize}
		\item $p(1,2)<p(2,1)$
		\item $c(3,3)<h_p(1,1)$
		\item $h_p(2,1)<\bigl(\,c(1,1), c(2,3)\,\bigr)$
		\item $\bigl(\,c(1,1), c(2,3)\,\bigr)<\Bigl(\,\bigl(\,p(1,1)\,\bigr)\,\Bigr)$
		\item $\bigl(\,c(1,1), c(2,3)\,\bigr)<\bigl(\,c(2,1), c(2,3)\,\bigr)$
	\end{itemize}
	\emptyline
	are all true.
\end{example}
\emptyline
As an implicit understanding, all sequences appearing In this study (except for the board solution mentioned later \cref{definition:BoardSolution}) are assumed to be in lexicographic ascending order (meaning that sequences with the same elements are uniquely determined and always sorted to be in the minimal lexicographic order). Therefore, note that all general set operations can be performed. For
example, In this study, when there are sequences $A$ and $S$, we describe "S is a subsequence of A" using the operator $\subset$ as in set theory, $S \subset A$. At this time, the operation $\mathfrak{P}{\mathrm{seq}}(A)$ of taking the power sequence of $A$ can be defined as
\begin{equation}
	\mathfrak{P}{\mathrm{seq}}(A)=(\,S\mid S\subset A\,)
\end{equation}
Due to the previous understanding, the order of the power sequence is assumed to be in lexicographic ascending order of $A$.

Here, we redefine the universal set dealt with In this study. For this purpose, we define an operation to flatten sequences as
\begin{equation}
	\text{flatten}(S) = \bigcup_{s \in S} \begin{cases}
		{s }              & \text{if } s \text{ is an element} \\
		\text{flatten}(s) & \text{if } s \text{ is a sequence}
	\end{cases}
\end{equation}
where $S$ is any sequence. At this time, using the previously mentioned $\mathbb{E}$, we define the universal set $\mathbb{U}$ as
\begin{equation}
	\mathbb{U} = \{\, A \mid \forall x \in A, \text{flatten}(x) \subseteq \text{flatten}(\mathbb{E}) \,\}
\end{equation}
Note that $\mathbb{U}$ cannot be identified with the infinite power set $\mathfrak{P}^\infty\bigl(\,\text{flatten}(\mathbb{E})\,\bigr)$ of $\text{flatten}(\mathbb{E})$.

Furthermore, we redefine the positional relationships of structures (vertical adjacency, horizontal adjacency, diagonal adjacency, coincidence) as follows.
\emptyline
\begin{definition}[Positional Relationships of Structures]\label{definition:RePositionalRelationship}
	When there are structures $X$ and $Y$, we define the positional relationship $\mathrm{R}$ as
	\begin{equation}
		\mathrm{R}(X,Y)\coloneqq \exists x \in X, \exists y \in Y, \mathrm{R}(x,y)
	\end{equation}
	where $\mathrm{R}\in \{\,\mathrm{H},\mathrm{V},\mathrm{D},\mathrm{M}\,\}$. Note that this definition is recursive.
\end{definition}
\emptyline
Here, we define what it means for a graph to be connected. Let
$\mathbf{R}=\{\,\mathrm{H},\mathrm{V},\mathrm{D},\mathrm{M}\,\}$ be the set composed of positional relationships. At this time, when there exists a sequence $S$ and we introduce an element $R$ of $\mathfrak{P}(\mathbf{R})$, we can define an edge sequence $E(S,R)$ by viewing $S$ as a vertex sequence. Also, we can consider a graph $G\bigl(\,S,E(S,R)\,\bigr)$ from this.
\begin{equation}
	E(S,R)=\bigl(\,(\,si,s_j\,)\in S\times S\mid r\in \forall R, r(\,s_i,s_j\,)\,\bigr)
\end{equation}
Furthermore, we define that a graph $G(S,E(S,R))$ is connected as
\begin{equation}
	\mathrm{con}\Bigl(\,G\bigl(\,S,E(S,R)\,\bigr)\,\Bigr)=\bigl[\,\forall u,v\in E(S,R), \exists P=(\,u=w_0,w_1,\dots ,w_k=v\,)\,\bigr]
\end{equation}
where the path $P$ satisfies the following conditions:
\begin{enumerate}
	\item $w_0=u \land w_k=v$
	\item $\forall i \in [0,k-1] \cap \mathbb{N}, (w_i,w{i+1}) \in E(S,R)$
\end{enumerate}
Also, we define that a graph $G'\bigl(\,S',E(S',R')\,\bigr)$ is a subgraph of graph $G\bigl(\,S,E(S,R)\,\bigr)$ as
\begin{equation}
	\mathrm{subgraph}(G',G)=\bigl[\,[\,S' \subseteq S\,]\land [\,E(S',R)\subseteq E(S,R)\,] \land[\, \forall(u,v) \in E(S',R), u,v\in S'\,] \,\bigr]
\end{equation}

Based on these, we define structures in pencil puzzles below. However, if we adopt the definition that "a structure is any element included in the universal set $\mathbb{U}$," we would create disorderly objects that are far from existing puzzle rules and unrecognizable to humans. Since pencil puzzles should be solvable by humans, we provide a definition that is easily recognizable to humans. For this purpose, using the aforementioned adjacency relationships, we define composition operations as a progressive computational method to create structures from elements.
\emptyline
\begin{definition}[Composition Operation]\label{definition:CompositionCalculation}
	Let $\mathbf{R}=\{\,\mathrm{H},\mathrm{V},\mathrm{D},\mathrm{M}\,\}$ be the set composed of positional relationships. At this
	time, when taking an element $R$ of $\mathfrak{P}(\mathbf{R})$ and an element $S$ of the structure sequence $\mathbb{S}$ at that point (definition to be given later), we define the operation

	\begin{alignat}{2}
		 & \mathrm{combine}(R,E) \dcomma& = \dcomma & (S\,|C_\textrm{o} \land \,S_\textrm{g}\,)                                                           \\
		 & C_\textrm{o}          \dcomma& = \dcomma & \mathrm{con}\Bigl(\,G'\bigl(\,S', E(S', R)\,\bigr)\,\Bigr)                                          \\
		 & S_\textrm{g}          \dcomma& = \dcomma & \mathrm{subgraph}\Bigl(\,G'\bigl(\,S', E(S', R), G(S, E(S, R))\,\bigr)\,\Bigr)
	\end{alignat}
	as the \textbf{composition operation}. Note that the very first structure sequence is $\mathbb{E}$ (\cref{definition:Elements}).

	We call the sequence resulting from this $\mathrm{combine}$ operation, or an element of $\mathbb{E}$, a \textbf{structure}, and define the \textbf{structure sequence} $\mathbb{S}'$ as $\mathbb{S}'=\mathbb{S}\cup(\,\mathbf{X}\,)$, where $\mathbf{X}$ is the structure created by this composition operation and newly added as an element of $\mathbb{S}$. What can be added as input for the next composition operation is an element of $\mathbb{S}'$. In this sense, it is progressive, and when considering the composition operation as a mapping, note that the domain differs with each composition operation.

	Intuitively, the composition operation is a sequence composed of partial connected graphs of the graph spanned when introducing the positional relationship $R$ to the structure $E$. Using these, we can create most of the structures in existing pencil puzzles.

	We will call the structure sequence $\mathbb{S}_{end}$ remaining after a finite number of composition operations the \textbf{structure sequence $\mathbb{S}_{end}$ possessed by a certain puzzle rule}. Note that $\mathbb{E} \subset \mathbb{S}_{end}$.

	Since simply stating "structure" can lead to confusion between "a set containing all of a certain structure" or "a
	specific structure existing on the board," we will refer to the former as "structure" and the latter as "the structure" (the result of $\mathrm{combine}$ is the former).
\end{definition}

\emptyline

\begin{example}[Structures Possessed by Slitherlink]\label{example:CompositionCalculation}
	After performing the composition operation
	\begin{alignat}{2}
		 & R            & = \dcomma & \{\,\mathrm{H},\mathrm{V},\mathrm{D}\,\}    \\
		 & E            & = \dcomma & \mathbf{Ep}\quad (\in \mathbb{E})           \\
		 & \mathbf{X_1} & = \dcomma & \mathrm{combine} (R,E) \label{equation:X_1}
	\end{alignat}
	the remaining structure sequence $\mathbb{S}=\mathbb{E}\cup(\,\mathbf{X_1}\,)$ is the structure sequence possessed by Slitherlink, and $\mathbf{X_1}$ is the structure possessed by Slitherlink. Note that in the case of Slitherlink, there is a condition that the structure is a closed curve and there is only one on the board, but this is restricted by the \textit{constraints} to be described later.
\end{example}
\emptyline
When a certain puzzle rule exists and it is determined what structures it possesses, we can define a board. A board is specified by individual puzzle rules and the size of the two-dimensional grid. In other words, we define a board as the union of elements that always exist and a subset of structures created by composition operations. Unless otherwise specified, we assume the size of the two-dimensional grid is $m\times n$ ($m,n\in\mathbb{N}$).
\emptyline
\begin{definition}
	Let there be a puzzle rule P, and let it possess structures $\mathbf{X_1},\mathbf{X_2},\dots$. We define that $B$ is a board of P if the following proposition is true:
	\begin{equation}
		B\in \bigl(\,\mathbb{E}\cup (\,X_1\,) \cup (\,X_2\,) \cup \dots \mid X_1 \in \mathfrak{P}{\mathrm{seq}}(\mathbf{X1}), X_2 \in \mathfrak{P}{\mathrm{seq}}(\mathbf{X_2}), \dots, \,\bigr)
	\end{equation}
\end{definition}
\emptyline
In pencil puzzles, structures existing on the board have corresponding states. We define the specific numerical values, constants, or vectors as \textbf{solutions} (e.g., 3, $x_1$,
(2, 3)). For convenience In this study, we specifically correspond edges to 1 when "effective" and 0 when "ineffective". Depending on the puzzle rule, there may be no corresponding state for a certain structure, in which case we specially define the value as \textbf{\textit{null}}.

From the above discussion, to define a puzzle rule, we need to define which set the solutions corresponding to the structures possessed by the puzzle rule belong to. For this purpose, we define the set of states using the term \textit{domain} below.
Note that when there is a certain board, not all structures contained in it necessarily have only one solution.
\emptyline
\begin{definition}[\textit{domain}]\label{definition:Domain}
	We define the set of possible states corresponding to structures possessed by a certain puzzle rule as the \textbf{\textit{domain}}. When a puzzle rule P possesses a structure $\mathbf{X_1}$, we will describe it in the form $\mathbf{X}\leftrightarrow S$ to mean the \textit{domain S} of $\mathbf{X_1}$.
\end{definition}
\emptyline
\begin{example}[Slitherlink's \textit{domain}]\label{example:SlitherLinkCodomain}
	The \textit{domains} of the structures possessed by Slitherlink are described as:
	\begin{alignat}{2}
		 & \mathbf{P}   & \leftrightarrow \dcomma & \{\,null\,\}      \\
		 & \mathbf{C}   & \leftrightarrow \dcomma & \{\,0,1,2,3,4\,\} \\
		 & \mathbf{E_p} & \leftrightarrow \dcomma & \{\,0,1\,\}       \\
		 & \mathbf{E_c} & \leftrightarrow \dcomma & \{\,null\,\}      \\
		 & \mathbf{X_1} & \leftrightarrow \dcomma & \{\,null\,\}
	\end{alignat}
	where $\mathbf{X_1}$ is given by \textup{\cref{equation:X_1}}.
\end{example}
\emptyline
By defining the \textit{domain}, when a certain board $B$ exists, we can consider possible solutions for each structure existing in $B$. If we describe the solution corresponding to a certain structure $X$ as $\mathrm{solution}(X)$, when viewed as a sequence (defined as a board solution), since the \textit{domain}
of the structure is not necessarily a singleton, we obtain a sequence of board solutions corresponding to $B$. We define this as the board solution sequence $\mathrm{S}(B)$.
\emptyline
\begin{definition}[Board Solution, Board Solution Sequence $\mathrm{S}(B)$]\label{definition:BoardSolution}
	When a certain board $B$ exists, we define the sequence of solutions corresponding to each structure contained in $B$ as the \textbf{board solution}. Also, we define the sequence composed of all board solutions in $B$ as the \textbf{board solution sequence} $\mathrm{S}(B)$ of board $B$. Note that among the sequences dealt with In this study, only the board solution is not in lexicographic ascending order, but corresponds one-to-one with the elements of board $B$.
\end{definition}
\emptyline
\begin{remark}
	When $B$ is a board,
	\begin{alignat}{2}
		 & \forall S_i \in \mathrm{S}(B)       & ,\dcomma & s=|B|=|S|   \\
		 & \forall i \in [1,s] \cap \mathbb{N} & ,\dcomma & |B_i|=|S_i| \\
	\end{alignat}
	holds. \textup{(The elements of $B$ and $S(B)$ correspond one-to-one.)}
\end{remark}
\emptyline
By determining the structure sequence and \textit{domain} existing in puzzle rule \textit{P}, the sequence composed of all boards in \textit{P} and the elements of the corresponding board solution sequence are determined. We define these as the board sequence $\mathfrak{B}$ and the board sequence solution sequence $\mathrm{S}(\mathfrak{B})$ below, respectively.
\emptyline
\begin{definition}[Board Sequence $\mathfrak{B}$, Board Sequence Solution Sequence $\mathrm{S}(\mathfrak{B})$]\label{definition:BoardSequence}
	When a certain puzzle rule P exists, we define the sequence composed of all possible boards from the structures possessed by P as the \textbf{board sequence} $\mathfrak{B}$. Also, we define the sequence of board solution sequences corresponding to each element of the board sequence as the \textbf{board sequence solution sequence} $\mathrm{S}(\mathfrak{B})$.
\end{definition}
\emptyline
\begin{remark}
	$|\mathrm{S}(\mathfrak{B})|=|\mathfrak{B}|$ holds \textup{(The elements of $\mathrm{S}(\mathfrak{B})$ and $\mathfrak{B}$ correspond one-to-one.).}
\end{remark}
\emptyline
\begin{example}
	For convenience, we consider the board sequence and board sequence solution sequence of Slitherlink on a very small size (2×2) two-dimensional grid. Note that since we are not considering \textit{constraints}, the board sequence includes boards that would not be possible in general Slitherlink.

	When
	\begin{alignat}{2}
		 & \mathbf{P}   & =\dcomma & \bigl(\,p(1,1),\dots p(3,3)\,\bigr)                                \\
		 & \mathbf{H_p} & =\dcomma & \bigl(\,h_p(1,1),\dots h_p(3,2)\,\bigr)                            \\
		 & \mathbf{V_p} & =\dcomma & \bigl(\,v_p(1,1),\dots v_p(2,3)\,\bigr)                            \\
		 & \mathbf{C}   & =\dcomma & \bigl(\,c(1,1),\dots c(2,2)\,\bigr)                                \\
		 & \mathbf{H_c} & =\dcomma & \bigl(\,h_c(1,1),h_c(2,1)\,\bigr)                                  \\
		 & \mathbf{V_c} & =\dcomma & \bigl(\,v_c(1,1),v_c(1,2)\,\bigr)                                  \\
		 & \mathbf{E_p} & =\dcomma & \bigl(\,h_p(1,1),\dots h_p(3,2), v_p(1,1),\dots v_p(2,3)\,\bigr)   \\
		 & \mathbf{E_c} & =\dcomma & \bigl(\,h_c(1,1),h_c(2,1),v_c(1,1),v_c(1,2)\,\bigr)                \\
		 & \mathbb{E}   & =\dcomma & \bigl(\,\mathbf{P} ,\mathbf{E_p}, \mathbf{C}, \mathbf{E_c}\,\bigr)
	\end{alignat}
	the board sequence $\mathfrak{B}$ becomes
	\begin{equation}
		\mathfrak{B}=\{\,\mathbb{E} \cup X_1 \mid X_1\in \mathfrak{P}(\mathbf{X_1})\,\}
	\end{equation}
	where $\mathbf{X_1}$ is given by \textup{\cref{equation:X_1}}. Note that $|\mathfrak{B}|=1$ does not hold. Due to space limitations, we omit writing out the entire board sequence solution sequence $\mathrm{S}(\mathfrak{B})$ corresponding to this board sequence $\mathfrak{B}$.
	When considering a board $B(\in \mathfrak{B})$ such as
	\begin{alignat}{2}
		B=\Bigl(\, & \bigl(\, p(1,1),\dots p(3,3)\,\bigr),                                        \\
		           & \bigl(\,c(1,1),\dots c(2,2)                \,\bigr)                          \\
		           & \bigl(\,h_p(1,1),\dots h_p(3,2), v_p(1,1),\dots v_p(2,3)            \,\bigr) \\
		           & \bigl(\,h_c(1,1),h_c(2,1),v_c(1,1),v_c(1,2)                  \,\bigr)        \\
		           & \bigl(\,(\, h(1,1),h(2,1),v(1,1),v(2,1)\,)\,\bigr)\,\Bigr)
	\end{alignat}
	a certain board solution $S\in \mathrm{S}(B)$ corresponding to the elements of this $B$ is expressed as
	\begin{alignat}{2}
		S=\Bigl(\, & \bigl(\, null,\dots ,null\,\bigr),                          \\
		           & \bigl(\,4,1,1,0              \,\bigr)                       \\
		           & \bigl(\,1,0,1,0,0,0,1,1,0,0,0,0           \,\bigr)          \\
		           & \bigl(\,null,null,null,null                  \,\bigr)       \\
		           & \bigl(\,null\,\bigr)\,\Bigr) \label{equation:BoardSolution}
	\end{alignat}
	Note that there are infinitely many elements included in $\mathrm{S}(B)$ other than \textup{\cref{equation:BoardSolution}}.
\end{example}

\emptyline
When dealing with puzzle rules, simply determining the structure sequence and \textit{domain} results in excessively large board sequences and board sequence solution sequences, as mentioned earlier. Also, in most existing puzzle rules, some restrictions are placed on structures existing on the board. We mathematically define these as \textit{constraints}. These essentially become a group of predicates that apply to board sequences and board sequence solution sequences.
\emptyline
\begin{definition}[\textit{constraints}]\label{definition:Constraints}
	We define a group of predicates with structures existing on the board and their corresponding solutions as free variables as \textbf{\textit{constraints}}. However, we require that the truth set of the logical product of this group of predicates is not an empty set. We defer the specific description method of \textit{constraints} to \textup{\cref{example:constraints}} described later.
\end{definition}
\emptyline
When \textit{constraints} are defined, their essence becomes a restriction mapping applied to board sequences and board sequence solution sequences. Here, we define board sequences and board sequence solution sequences restricted by \textit{constraints} as constrained board sequences and constrained board sequence solution sequences, respectively.
\emptyline
\begin{definition}[Constrained Board Sequence $\mathfrak{B}_{\mathrm{con}}$, Constrained Board Sequence Solution Sequence $\mathrm{S}(\mathfrak{B}_{\mathrm{con}})$]
	We define \textbf{constrained board sequence} $\mathfrak{B}_{\mathrm{con}}$ and \textbf{constrained board sequence solution sequence} $\mathrm{S}(\mathfrak{B}_{\mathrm{con}})$ respectively as
	\begin{alignat}{2}
		 & \mathfrak{B}_{\mathrm{con}}            & \coloneqq \dcomma & \bigl(\,B\mid B\in \mathfrak{B}, \mathrm{S}(B)\in \mathrm{S}(\mathfrak{B}), \textit{constraints},(B,\mathrm{S}(B))\,\bigr)            \\
		 & \mathrm{S}(\mathfrak{B}_{\mathrm{con}}) & \coloneqq \dcomma & \bigl(\,\mathrm{S}(B)\mid B\in \mathfrak{B},\mathrm{S}(B)\in \mathrm{S}(\mathfrak{B}), \textit{constraints},(B,\mathrm{S}(B))\,\bigr)
	\end{alignat}
	These correspond to what are called "answers" or "solution boards" in existing pencil puzzles.
\end{definition}
\emptyline
\begin{remark}
	$|\mathrm{S}(\mathfrak{B}_{\mathrm{con}})|=|\mathfrak{B}_{\mathrm{con}}|$ holds. Also, $\mathfrak{B}_{\mathrm{con}}\subset \mathfrak{B}$ and $\mathrm{S}(\mathfrak{B}_{\mathrm{con}}) \subset \mathrm{S}(\mathfrak{B})$ hold.
\end{remark}
\emptyline
\begin{example}\label{example:constraints}
	The \textit{constraints} of Slitherlink are described as follows. Note that we are using a group of functions that can be used in \textit{constraints} as described in.
	\emptyline
	\begin{itemize}
		\item $\forall X_1 \in \mathbf{X_1}, \forall x_1 \in X_1, \mathrm{solution}(x_1)=1$
		\item $\forall p \in \mathrm{B}(\mathbf{P}), \mathrm{cross}(p)=2$
		\item $\forall c \in \mathrm{B}(\mathbf{C}), \mathrm{solution}(c)=\mathrm{cycle}(c)$
		\item $|\mathrm{B}(\mathbf{X_1})|=1$
	\end{itemize}
	\emptyline
	Each bullet point becomes a group of predicates that restrict the free variables of board sequences and board sequence solution sequences with the expression
	\begin{alignat}{2}
		 & \bigl[\,\forall X_1 \in \mathbf{X_1}, \forall x_1 \in X_1, \mathrm{solution}(x_1)=1\,\bigr]\land \\
		 & \bigl[\,\forall p \in B(\mathbf{P}), \mathrm{cross}(p)=2\,\bigr]\land                            \\
		 & \bigl[\,\forall c \in B(\mathbf{C}), \mathrm{solution}(c)=\mathrm{cycle}(c)\,\bigr]\land         \\
		 & \bigl[\,|B(\mathbf{X_1})|=1\,\bigr]
	\end{alignat}
	We used the above expression for simplicity.
\end{example}
\emptyline

Furthermore, we also define the act of "presenting a problem". As a preliminary step, we define the constant \textit{undecided} and the operator of extended subsequence $\subset{\mathrm{ext}}$.
\begin{definition}[\textit{undecided}, $\subset{\mathrm{ext}}$]
	We define that a sequence $A$ \textbf{contains sequence $B$ as an extended subsequence} ($B \subset{\mathrm{ext}} A$) if "for any element $bi$ of $B$, there exists a corresponding element $a_i$ in $A$, and $b_i=a_i$ or $b_i=undecided$." This can be formally described as
	\begin{equation}
		B\subset{\mathrm{ext}} A \iff \forall i ,1\leq i \leq |B|, bi\in B, a_i \in A, [\,b_i=a_i\,] \lor [\,b_i=undecided\,]
	\end{equation}
	Unlike general subsequences,
	\textup{
		\emptyline
		\begin{enumerate}
			\item Elements must match from the beginning.
			\item No comparison is made in the case of \textit{undecided}.
		\end{enumerate}
		\emptyline
		Note these points.
	}
\end{definition}
\emptyline
\begin{definition}[Presentation of a Problem]\label{definition:PresentQuiz}
	When there exists a puzzle rule $P$, we define $\mathfrak{Q}$ and $S(\mathfrak{Q})$ as
	\begin{equation}
		\mathfrak{Q}\subset \mathfrak{B}, S(\mathfrak{Q})=\Bigl(\,\bigl(\,Q\mid Q\in S(B), S_Q\subset{\mathrm{ext}} Q\,\bigr)\mid S(B)\in S(\mathfrak{B})\,\Bigr)
	\end{equation}
	where $SQ$ is an arbitrary sequence (however, it actually follows \cref{definition:hidden} described later). We define \textbf{presentation of a problem} as presenting $\mathfrak{Q}$ and $S_Q$ to the solver (human or machine solving the puzzle) such that
	\begin{alignat}{2}
		 & s=|S(\mathfrak{B}\mathrm{con})|, Q_i\in \mathfrak{Q}, S(B)i\in S(\mathfrak{B}\mathrm{con}) \\
		 & \exists! i\in [1,s]\cap \mathbb{N}, |Q_i\cap S(B)_i|=1
	\end{alignat}
	holds. Also, we define the combination of $[\,\mathfrak{Q},S_Q\,]$ as a \textbf{problem}.
\end{definition}
\emptyline
Existing pencil puzzles are implicitly required to have a unique solution. Therefore, we adopted the definition like \cref{definition:PresentQuiz}. By determining the structure sequence, \textit{domain}, and \textit{constraints}, the constrained board sequence and
constrained board sequence solution sequence are determined. However, the existence of $\mathfrak{Q}$ and $S_Q$ is not self-evident. The existence of $\mathfrak{Q}$ and $S_Q$ is a necessary condition for being a puzzle rule. Currently, there is no method other than relying on computational methods to examine whether $\mathfrak{Q}$ and $S_Q$ exist from the combination of structure sequence, \textit{domain}, and \textit{constraints}, so this remains a future task.
\emptyline
\begin{remark}[Necessary Condition for Being a Puzzle Rule]\label{remark:OnlySolution}
	The existence of $\mathfrak{Q}$ and $S_Q$ satisfying \textup{\cref{definition:PresentQuiz}} is a necessary condition for being a puzzle rule.
\end{remark}
\emptyline

Although $S_Q$ is an arbitrary sequence, the index for choosing it depends on the puzzle rule (in Slitherlink, $S_Q$ is presented with all solutions corresponding to grid point vertical edges and grid point horizontal edges included in the board, and all cells set to \textit{undecided}). Generally, since some solutions corresponding to structures possessed by a certain puzzle rule are often presented as \textit{undecided}, we define this information as \textit{hidden}.
\emptyline

\begin{definition}[\textit{hidden}]\label{definition:hidden}
	When there exists a puzzle rule P, the range of solutions corresponding to a structure $X$ possessed by P is specified by the \textit{domain}. When there exists a board $B$, the board solution is composed of solutions determined within the range of that \textit{domain}, but we construct a new $S_Q$ from a board solution with this \textit{domain} modified. We define \textbf{\textit{hidden}} as which structure's \textit{domain} to add \textit{undecided} to. We defer the specific description method to \textup{\cref{example:hidden}}.
\end{definition}
\emptyline

\begin{example}[Slitherlink's \textit{hidden}]\label{example:hidden}
	Slitherlink's \textit{hidden} is described as follows. Note that being
	\textit{hidden} of a certain structure $\mathbf{X_1}$ is described as $\mathbf{X_1}\rightarrow S'$, meaning "\textit{hidden S'} of $\mathbf{X_1}$", similar to the description of \textit{domain}.

	\begin{alignat}{4}
		 & \mathbf{P}   & \leftrightarrow \dcomma & \{\,null\,\}     \dcomma & \rightarrow \dcomma & \{\,null\,\}                \\
		 & \mathbf{C}   & \leftrightarrow \dcomma & \{\,0,1,2,3,4\,\}\dcomma & \rightarrow \dcomma & \{\,0,1,2,3,4,undecided\,\} \\
		 & \mathbf{E_p} & \leftrightarrow \dcomma & \{\,0,1\,\}      \dcomma & \rightarrow \dcomma & \{\,undecided\,\}           \\
		 & \mathbf{E_c} & \leftrightarrow \dcomma & \{\,null\,\}     \dcomma & \rightarrow \dcomma & \{\,null\,\}                \\
		 & \mathbf{X_1} & \leftrightarrow \dcomma & \{\,null\,\}     \dcomma & \rightarrow \dcomma & \{\,null\,\}
	\end{alignat}
	This is \textup{\cref{example:SlitherLinkCodomain}} with \textit{hidden} added. For example, for grid point edges, \textit{hidden} becomes a singleton consisting only of \textit{undecided}. In practice, $S_Q$ can be constructed as if \textit{hidden} were the \textit{domain}.
\end{example}
\emptyline
Thus, a puzzle rule can be defined by determining the structure sequence $\mathbb{S}$, \textit{domain}, \textit{constraints}, and \textit{hidden}.

\emptyline

\begin{definition}[Puzzle Rule]\label{definition:PuzzleRule}
	We define a \textbf{puzzle rule} as a combination of structure sequence $\mathbb{S}$, \textit{domain}, \textit{constraints}, and \textit{hidden} that satisfies \textup{\cref{remark:OnlySolution}}.
\end{definition}
\emptyline
Using this, in \cref{section:verification}, we evaluate whether existing puzzle rules can be described using the puzzle rules of this research.

\section{Verification of Puzzle Rules}\label{section:verification}
In this section, we experiment with describing existing puzzle rules using the mathematical expressions defined In this study and verify their effectiveness through computational implementation. The existing puzzle rules we dealt with are the Nikoli Puzzles  posted on \citet{Nikoli2024}, Ltd. website. We computationally implemented the existing puzzle rules described using the mathematical expressions defined In this study, verified that the completed boards output from them did not differ from the existing puzzle rules, and thereby corroborated that the definitions In this study are to some extent valid.

\subsection{Method}\label{subsection:method}
The method for verifying effectiveness was carried out in the following three steps. The target puzzle rules are the Nikoli Puzzles posted on \citet{Nikoli2024}, Ltd. website.

\begin{enumerate}
	\item Converted existing puzzle rules into mathematical expressions following \cref{definition:PuzzleRule}.
	\item Implemented the mathematical expressions converted in 1. computationally. Specifically, we implemented the generation of completed boards according to the mathematical expressions.
	\item Confirmed from the output whether the completed boards created using 2. do not deviate from the completed boards of existing puzzle rules. If there is no deviation here, it suggests that at least for that mathematical expression, it is a sufficient
	      condition for the existing puzzle rule.
\end{enumerate}

Note that what is verified here is only a sufficient condition, and we cannot confirm the necessary condition.

\subsection{Results}\label{subsection:result}
Out of a total of 46, 10 were confirmed to be sufficient conditions by the method described in \cref{subsection:method}. Specifically, we confirmed that those listed in \cref{table:OKPuzzles} are sufficient conditions (including "Inshi no heya" and "Sukoro" which are not posted in \citet{Nikoli2024}. There are 44 puzzles posted in \citet{Nikoli2024}.). The mathematical expressions are posted in \cref{subsection:MathematicalExpression}, and the computational implementations are posted in \url{https://github.com/itkmaingit/puzzle_check}.

\begin{table}[ht]
	\caption{Puzzles confirmed to be sufficient conditions}
	\centering
	\begin{tabular}{ll}
		\toprule
		\multicolumn{1}{c}{Name}\\
		\midrule
		Choco Banana     \\
		Kurotto      \\
		Fillomino        \\
		Inshi no heya \\
		Hitori           \\
		Sudoku        \\
		Sukoro           \\
		Norinori      \\
		Shikaku          \\
		Slitherlink   \\
		\bottomrule
	\end{tabular}
	\label{table:OKPuzzles}
\end{table}

\subsection{Discussion}\label{subsection:consdiration}
Those that were not confirmed were classified as puzzle rules falling under any of the following categories, or (even if successful in mathematical expression) difficult to implement computationally.

\begin{enumerate}
	\item The graph contains directional information (directed graph).
	\item Elements that cannot be expressed by grid points, cells, grid point edges, or cell edges are included.
	\item Information exists outside the board. \label{enumerate:outBoard}
	\item It has structures that cannot be created by composition operations.
\end{enumerate}

For example, Nansuke falls under the limitation in 3.

\section{Conclusion}\label{section:Conclusion}
In this study, we mathematically defined puzzle rules for pencil puzzles. We demonstrated its capacity to express existing puzzle rules, successfully representing one-fourth of the analyzed puzzle rules. In addition, we identified cases where certain puzzle rules could not be expressed due to an inherent limitation of the definition or computational challenges. This study highlights the potential for creating new puzzle rules by replacing or combining mathematical expressions of existing puzzle rules. In particular, when the structure spaces would match, the identical heuristic functions can be applied, making it possible to combine constraints. This combining expression, in turn, enables systematic rule creation, broadening the potential applications that range from enhancing engagement in recreational puzzles to advancing logic-based systems in educational and computational contexts. Ultimately, this study lays a robust foundation for automating puzzle rule generation, supporting both theoretical exploration and practical implementation.

\section{Acknowledgement}\label{section:acknowledgement}
The contributions of I.M. included conceptualization, investigation, data curation, methodology, validation, and drafting the initial version of this manuscript. Y.I. contributed to the conceptualization, funding acquisition, supervision, and the review and editing of this manuscript.

\bibliographystyle{unsrtnat}
% \bibliography{references}  %%% Uncomment this line and comment out the

\appendix
\renewcommand{\thesection}{\Alph{section}}
\renewcommand{\thesubsection}{\Alph{section}.\arabic{subsection}}
\setcounter{section}{0}
\renewcommand{\theequation}{\Alph{section}.\arabic{equation} }
\setcounter{equation}{0}
\section{Appendix}\label{section:Appendix}
\subsection{Heuristic Function Group}\label{subsection:HeuristicFunctions}
Below are the operations used in \textit{constraints}, or predicates and their correspondences.
\subsubsection{\textrm{B}}
A function that takes a structure $\mathbf{X}$ possessed by a puzzle rule as an argument and returns the structures included in the board $\mathrm{B}$.
\begin{equation}
	\begin{array}{rccc}
		\mathrm{B}\colon & \mathrm{B}(\mathbf{X}) & \longmapsto & \{\,X\mid X\in B \cap   \mathbf{X}\,\}.
	\end{array}
\end{equation}
\subsubsection{\textrm{cross}}
A function that calculates the number of grid point edges gathering at a certain grid
point. When the coordinates of the grid point are $(\,i,j\,)$, with $E=\{\,hp(i,j-1),h_p(i,j),v_p(i-1,j),v_p(i,j)\,\}$,
\begin{equation}
	\begin{array}{rccc}
		\mathrm{cross}\colon & \mathbf{P} & \longrightarrow & \mathbb{Z}                                                                                  \\
		                     & p(i,j)     & \longmapsto     & \sum\limits{e(k, l)\in E}\begin{cases}
			                                                                               \mathrm{solution}(e(k,l)) & \text{if } e(k,l) \in \mathbf{Ep},     \\
			                                                                               0                         & \text{if } e(k,l) \notin \mathbf{E_p}.
		                                                                               \end{cases}
	\end{array}
\end{equation}
\subsubsection{\textrm{cycle}}
A function that calculates the number of grid point edges gathering around a certain cell. When the coordinates of the cell are $(\,i,j\,)$, with $E=\{\,h_p(i,j),h_p(i+1,j),v_p(i,j),v_p(i,j+1)\,\}$,
\begin{equation}
	\begin{array}{rccc}
		\mathrm{cross}\colon & \mathbf{C} & \longrightarrow & \mathbb{Z}                                                                                  \\
		                     & c(i,j)     & \longmapsto     & \sum\limits{e(k, l)\in E}\begin{cases}
			                                                                               \mathrm{solution}(e(k,l)) & \text{if } e(k,l) \in \mathbf{E_p},    \\
			                                                                               0                         & \text{if } e(k,l) \notin \mathbf{E_p}.
		                                                                               \end{cases}
	\end{array}
\end{equation}
\subsubsection{\textrm{all\_different}}
A predicate that becomes true when the solutions of the structures included in the structure given as an argument are all different.
\begin{equation}
	\begin{array}{rccc}
		\mathrm{all\_different}(X)=\bigl[\,\forall x, y \in X, , (x \neq y \Rightarrow f(x) \neq f(y))\,\bigr].
	\end{array}
\end{equation}
\subsubsection{\textrm{is\_rectangle}}
A predicate that becomes true when the structure given as an argument forms a rectangle when arranged on the board. However, the structures that can be given as arguments are only elements included in the structure sequence that has undergone composition operation only once.
\begin{alignat}{2}
	\mathrm{is\_rectangle}(X)= & \bigl[\,\forall x(i,j) \in X, , \exists i, j \in \mathbb{Z}, (\text{{minX}} \leq i \leq \text{{maxX}} \land \text{{minY}} \leq j \leq \text{{maxY}})\,\bigr]
	\\ \land  &\bigl[\, (\text{{maxX}} - \text{{minX}}) \times (\text{{maxY}} - \text{{minY}}) = |X|\,\bigr].
\end{alignat}
Here, minX, maxX, minY, maxY are the minimum and maximum values of the coordinates of the elements included in structure X.
\subsubsection{\textrm{is\_square}}
A predicate that becomes true when the structure given as an argument forms a square when arranged on the board. However, the structures that can be given as arguments are only elements included in the structure sequence that has undergone composition operation only once.
\begin{alignat}{2}
	\mathrm{is\_square}(X)= & \bigl[\,\mathrm{is\_rectangle}(X)\,\bigr]
	\\ \land &\bigl[\, (\text{{maxX}} - \text{{minX}}) \times (\text{{maxY}} - \text{{minY}}) = |X|\,\bigr].
\end{alignat}
Here, minX, maxX, minY, maxY are the minimum and maximum values of the coordinates of the elements included in structure X.
\subsubsection{\textrm{connect}}
When a board $B$ exists, a function that returns structures connected by the positional relationship given as an argument to the structure given as an argument. With $R\subset \mathbf{R}=\{\,\mathrm{H},\mathrm{V},\mathrm{D},\mathrm{M}\,\}$,
\begin{equation}
	\begin{array}{rccc}
		\mathrm{connect}\colon & \mathrm{connect}(X, R) & \longmapsto \{\,X{other}\mid(X,X{other})\in
		E(\mathrm{B}(\mathbf{X},R))\,\}.
	\end{array}
\end{equation}
\subsubsection{\textrm{no\_overlap}}
When given a structure $\mathbf{X}$ possessed by a puzzle rule, a predicate that becomes true when the intersection is an empty set when the flattening function $\mathrm{flatten}$ is applied to all structures belonging to it within a certain board $B$. When multiple structures are given, it is to construct a union within the board $B$.
\begin{equation}
	\mathrm{no\_overlap}(\mathbf{X}) =\bigl[\,\forall X_1, X_2\in \mathrm{B}(\mathbf{X}), (\,X_1 \neq X_2 \Rightarrow \mathrm{flatten}(X_1) \cap \mathrm{flatten}(X_2)=\emptyset \,)\,\bigr]
\end{equation}
\subsubsection{\textrm{fill}}
When given a structure $\mathbf{X}$ possessed by a puzzle rule, a predicate that becomes true when, within a certain board $B$, when the flattening function $\mathrm{flatten}$ is applied to that structure, it matches one of the element sequences and satisfies the $\mathrm{overlap}$ function.
\begin{alignat}{2}
	\mathrm{fill}(\mathbf{X})= & \bigl[\,\mathrm{no\_overlap}(\mathbf{X})\,\bigr] \land \bigl[\,\mathrm{flatten}(\mathrm{B}(\mathbf{X})) \in \mathbb{E}\,\bigr].
\end{alignat}

\subsection{Conversion of Existing Puzzle Rules to Mathematical Expressions}\label{subsection:MathematicalExpression}
Below are the existing puzzle rules converted to mathematical expressions as demonstrated by \cref{section:verification}. However, we first describe commonly used structures.
\begin{alignat}{2}
	 & R          & = \dcomma & \{\,\mathrm{H},\mathrm{V}\,\}                \\
	 & E          & = \dcomma & \mathbf{C}\quad (\in \mathbb{E})             \\
	 & \mathbf{A} & = \dcomma & \mathrm{combine} (R,E) \label{equation:area}
\end{alignat}

\begin{alignat}{2}
	 & R            & = \dcomma & \{\,\mathrm{H}\,\}                            \\
	 & E            & = \dcomma & \mathbf{C}\quad (\in \mathbb{E})              \\
	 & \mathbf{A_h} & = \dcomma & \mathrm{combine} (R,E)\label{equation:area_h}
\end{alignat}

\begin{alignat}{2}
	 & R            & = \dcomma & \{\,\mathrm{V}\,\}                            \\
	 & E            & = \dcomma & \mathbf{C}\quad (\in \mathbb{E})              \\
	 & \mathbf{A_v} & = \dcomma & \mathrm{combine} (R,E)\label{equation:area_v}
\end{alignat}

\begin{alignat}{2}
	 & R            & = \dcomma & \{\,\mathrm{H},\mathrm{V},\mathrm{D}\,\}        \\
	 & E            & = \dcomma & \mathbf{E_p}\quad (\in \mathbb{E})              \\
	 & \mathbf{G_p} & = \dcomma & \mathrm{combine} (R,E) \label{equation:graph_p}
\end{alignat}
\begin{alignat}{2}
	 & R            & = \dcomma & \{\,\mathrm{H},\mathrm{V},\mathrm{D}\,\}         \\
	 & E            & = \dcomma & \mathbf{E_c}\quad (\in \mathbb{E})               \\
	 & \mathbf{G_c} & = \dcomma & \mathrm{combine} (R,E). \label{equation:graph_c}
\end{alignat}

\subsubsection{Slitherlink}\label{subsubsection:slitherlink}

Possessed structure is $\mathbf{G_p}$. \textit{domain} and \textit{hidden} are
\begin{alignat}{4}
	 & \mathbf{P}   & \leftrightarrow \dcomma & \{\,null\,\}     \dcomma & \rightarrow \dcomma & \{\,null\,\}
	\\
	 & \mathbf{C}   & \leftrightarrow \dcomma & \{\,0,1,2,3,4\,\}\dcomma & \rightarrow \dcomma & \{\,0,1,2,3,4,undecided\,\} \\
	 & \mathbf{E_p} & \leftrightarrow \dcomma & \{\,0,1\,\}      \dcomma & \rightarrow \dcomma & \{\,undecided\,\}           \\
	 & \mathbf{E_c} & \leftrightarrow \dcomma & \{\,null\,\}     \dcomma & \rightarrow \dcomma & \{\,null\,\}                \\
	 & \mathbf{G_p} & \leftrightarrow \dcomma & \{\,null\,\}     \dcomma & \rightarrow \dcomma & \{\,null\,\}.
\end{alignat}

\textit{constraints} are
\begin{alignat}{2}
	 & \forall G_p \in \mathbf{G_p}, \forall e_p \in G_p, \mathrm{solution}(e_p)=1  & \\
	 & \forall p \in \mathrm{B}(\mathbf{P}), \mathrm{cross}(p)=2                    & \\
	 & \forall c \in \mathrm{B}(\mathbf{C}), \mathrm{solution}(c)=\mathrm{cycle}(c) & \\
	 & |\mathrm{B}(\mathbf{G_p})|=1.                                                &
\end{alignat}

\subsubsection{Sudoku}\label{subsubsection:sudoku}

Possessed structures are $\mathbf{A_h}$, $\mathbf{A_v}$, $\mathbf{A}$. \textit{domain} and \textit{hidden} are
\begin{alignat}{4}
	        & \mathbf{P}               & \leftrightarrow \dcomma & \{\,null\,\}     \dcomma         & \rightarrow \dcomma & \{\,null\,\}                        \\
	        & \mathbf{C}               & \leftrightarrow \dcomma & \{\,1,2,3,4,5,6,7,8,9\,\}\dcomma & \rightarrow \dcomma & \{\,1,2,3,4,5,6,7,8,9,undecided\,\} \\
	        & \mathbf{E_p}             & \leftrightarrow \dcomma & \{\,null\,\}      \dcomma        & \rightarrow \dcomma & \{\,null\,\}                        \\
	        & \mathbf{E_c}             & \leftrightarrow \dcomma & \{\,null\,\}     \dcomma         & \rightarrow \dcomma & \{\,null\,\}                        \\
	        & \mathbf{A_h}             & \leftrightarrow \dcomma & \{\,null\,\}     \dcomma         & \rightarrow \dcomma & \{\,null\,\}                        \\
	        & \mathbf{A_v}             & \leftrightarrow \dcomma & \{\,null\,\}     \dcomma         & \rightarrow \dcomma & \{\,null\,\}                        \\
	        & \mathbf{A}               & \leftrightarrow
	\dcomma & \{\,null\,\}     \dcomma & \rightarrow \dcomma     & \{\,null\,\}.
\end{alignat}

\textit{constraints} are
\begin{alignat}{2}
	 & \mathrm{fill}(\mathbf{A_h})                                                                                  & \\
	 & \mathrm{fill}(\mathbf{A_v})                                                                                  & \\
	 & \mathrm{fill}(\mathbf{A})                                                                                    & \\
	 & \forall A_h \in \mathrm{B}(\mathbf{A_h}), |A|=9 \land \mathrm{all\_different}(A_h)                           & \\
	 & \forall A_v \in \mathrm{B}(\mathbf{A_v}), |A|=9 \land \mathrm{all\_different}(A_v)                           & \\
	 & \forall A \in   \mathrm{B}(\mathbf{A}), \mathrm{is\_square}(A) \land |A|=9 \land  \mathrm{all\_different}(A) &
\end{alignat}

\subsubsection{Shikaku}\label{subsubsection:shikaku}

Possessed structure is $\mathbf{A}$. \textit{domain} and \textit{hidden} are
\begin{alignat}{4}
	 & \mathbf{P}   & \leftrightarrow \dcomma & \{\,null\,\}     \dcomma              & \rightarrow \dcomma & \{\,null\,\}               \\
	 & \mathbf{C}   & \leftrightarrow \dcomma & \{\,null\,\}\dcomma                   & \rightarrow \dcomma & \{\,null\,\}               \\
	 & \mathbf{E_p} & \leftrightarrow \dcomma & \{\,null\,\}      \dcomma             & \rightarrow \dcomma & \{\,null\,\}               \\
	 & \mathbf{E_c} & \leftrightarrow \dcomma & \{\,null\,\}     \dcomma              & \rightarrow \dcomma & \{\,null\,\}               \\
	 & \mathbf{A}   & \leftrightarrow \dcomma & \{\,1,\dots n\times m\,\}     \dcomma & \rightarrow \dcomma & \{\,1,\dots n\times m\,\}.
\end{alignat}

\textit{constraints} are
\begin{alignat}{2}
	 & \mathrm{fill}(\mathbf{A})                                                                         & \\
	 & \forall A \in \mathrm{B}(\mathbf{A}), \mathrm{is\_rectangle}(A) \land \mathrm{solution}(A) = |A|. &
\end{alignat}
\subsubsection{Choco Banana}\label{subsubsection:choco_banana}

Possessed structures are $\mathbf{A_1}$, $\mathbf{A_2}$ (both from \cref{equation:area}). \textit{domain} and \textit{hidden} are
\begin{alignat}{4}
	 & \mathbf{P}   & \leftrightarrow \dcomma & \{\,null\,\}     \dcomma              & \rightarrow \dcomma & \{\,null\,\}               \\
	 & \mathbf{C}   & \leftrightarrow \dcomma & \{\,null\,\}\dcomma                   & \rightarrow \dcomma & \{\,null\,\}               \\
	 & \mathbf{E_p} & \leftrightarrow \dcomma & \{\,null\,\}      \dcomma             & \rightarrow \dcomma & \{\,null\,\}               \\
	 & \mathbf{E_c} & \leftrightarrow \dcomma & \{\,null\,\}     \dcomma              & \rightarrow \dcomma & \{\,null\,\}               \\
	 & \mathbf{A_1} & \leftrightarrow \dcomma & \{\,1,\dots n\times m\,\}     \dcomma & \rightarrow \dcomma & \{\,1,\dots n\times m\,\}. \\
	 & \mathbf{A_2} & \leftrightarrow \dcomma & \{\,1,\dots n\times m\,\}     \dcomma & \rightarrow \dcomma & \{\,1,\dots n\times m\,\}.
\end{alignat}

\textit{constraints} are
\begin{alignat}{2}
	 & \mathrm{fill}(\mathbf{A_1}, \mathbf{A_2})                                                                         & \\
	 & \forall A_1 \in \mathrm{B}(\mathbf{A_1}), \lnot \mathrm{is\_rectangle}(A_1) \land \mathrm{solution}(A_1) = |A_1|. & \\
	 & \forall A_2 \in \mathrm{B}(\mathbf{A_2}), \mathrm{is\_rectangle}(A_2) \land \mathrm{solution}(A_2) = |A_2|.       &
\end{alignat}

\subsubsection{Inshi no Heya}\label{subsubsection:inshi_no_heya}

Possessed structures are $\mathbf{A_h}$, $\mathbf{A_v}$, $\mathbf{A}$. \textit{domain} and \textit{hidden} are
\begin{alignat}{4}
	        & \mathbf{P}          & \leftrightarrow \dcomma & \{\,null\,\}     \dcomma & \rightarrow \dcomma & \{\,null\,\}                 \\
	        & \mathbf{C}          & \leftrightarrow \dcomma & \{\,null\,\}\dcomma      & \rightarrow \dcomma & \{\,null\,\}                 \\
	        & \mathbf{E_p}        & \leftrightarrow \dcomma & \{\,null\,\}
	\dcomma & \rightarrow \dcomma & \{\,null\,\}                                                                                            \\
	        & \mathbf{Ec}         & \leftrightarrow \dcomma & \{\,null\,\}     \dcomma & \rightarrow \dcomma & \{\,null\,\}                 \\
	        & \mathbf{A_h}        & \leftrightarrow \dcomma & \{\,null\,\}     \dcomma & \rightarrow \dcomma & \{\,null\,\}                 \\
	        & \mathbf{A_v}        & \leftrightarrow \dcomma & \{\,null\,\}     \dcomma & \rightarrow \dcomma & \{\,null\,\}                 \\
	        & \mathbf{A}          & \leftrightarrow \dcomma & \{\,null\,\}     \dcomma & \rightarrow \dcomma & \{\,1,\dots, n\times n!\,\}.
\end{alignat}

\textit{constraints} are
\begin{alignat}{2}
	 & n=m                                                                                                                                    & \\
	 & \mathrm{fill}(\mathbf{A_h})                                                                                                            & \\
	 & \mathrm{fill}(\mathbf{A_v})                                                                                                            & \\
	 & \mathrm{fill}(\mathbf{A})                                                                                                              & \\
	 & \forall A_h \in \mathrm{B}(\mathbf{A_h}), |A|=9 \land \mathrm{all\_different}(A_h)                                                     & \\
	 & \forall A_v \in \mathrm{B}(\mathbf{A_v}), |A|=9 \land \mathrm{all\_different}(A_v)                                                     & \\
	 & \forall A \in \mathrm{B}(\mathbf{A}), \mathrm{is\_rectangle}(A) \land \mathrm{solution}(A) = \prod\limits{c\in A}\mathrm{solution}(c). &
\end{alignat}

\subsubsection{Fillomino}\label{subsubsection:fillomino}

Possessed structure is $\mathbf{A}$. \textit{domain} and \textit{hidden} are
\begin{alignat}{4}
	 & \mathbf{P}   & \leftrightarrow \dcomma & \{\,null\,\}     \dcomma              & \rightarrow \dcomma & \{\,null\,\}                         \\
	 & \mathbf{C}   & \leftrightarrow \dcomma & \{\,1,\dots n\times m\,\}\dcomma      & \rightarrow \dcomma & \{\,1,\dots n\times m,undecided\,\}  \\
	 & \mathbf{Ep}  & \leftrightarrow \dcomma & \{\,null\,\}      \dcomma             & \rightarrow \dcomma & \{\,null\,\}                         \\
	 & \mathbf{E_c} & \leftrightarrow \dcomma & \{\,null\,\}     \dcomma              & \rightarrow \dcomma & \{\,null\,\}                         \\
	 & \mathbf{A}   & \leftrightarrow \dcomma & \{\,1,\dots n\times m\,\}     \dcomma & \rightarrow \dcomma & \{\,1,\dots n\times m,undecided\,\}.
\end{alignat}

\textit{constraints} are
\begin{alignat}{2}
	 & \mathrm{fill}(\mathbf{A})                                                                                       & \\
	 & \forall A \in \mathrm{B}(\mathbf{A}), \forall c \in A, \mathrm{solution}(A)=\mathrm{solution}(c)=|A|            & \\
	 & \forall A \in \mathrm{B}(\mathbf{A}),\forall A{other} \in \mathrm{connect}(A, \{\,H,V\,\}), |A|\neq |A_{other}| &
\end{alignat}

\subsubsection{Kurotto}\label{subsubsection:kurotto}

Possessed structure is $\mathbf{A}$. \textit{domain} and \textit{hidden} are
\begin{alignat}{4}
	 & \mathbf{P}   & \leftrightarrow \dcomma & \{\,null\,\}     \dcomma                  & \rightarrow \dcomma & \{\,null\,\}                               \\
	 & \mathbf{C}   & \leftrightarrow \dcomma & \{\,null,0,\dots n\times m-1,x\,\}\dcomma & \rightarrow \dcomma & \{\,null,0,\dots n\times m-1,undecided\,\} \\
	 & \mathbf{E_p} & \leftrightarrow \dcomma & \{\,null\,\}      \dcomma                 & \rightarrow \dcomma & \{\,null\,\}                               \\
	 & \mathbf{E_c} & \leftrightarrow \dcomma & \{\,null\,\}     \dcomma                  & \rightarrow \dcomma & \{\,null\,\}
	\\
	 & \mathbf{A}   & \leftrightarrow \dcomma & \{\,null\,\}     \dcomma                  & \rightarrow \dcomma & \{\,null\,\}.
\end{alignat}

\textit{constraints} are
\begin{alignat}{2}
	 & \mathrm{no\_overlap}(\mathbf{A})                                                                                                                           & \\
	 & \forall A \in \mathrm{B}(\mathbf{A}),\mathrm{connect}(A, \{\,H,V\,\})=\emptyset                                                                            & \\
	 & \forall c \in \mathrm{B}(\mathbf{C}), \mathrm{solution}(c)\neq null                                                                                          \\
	 & \Rightarrow \mathrm{solution}(c)=\sum\limits{c_{other}\in \mathrm{connect}(c,\{\,H,V\,\})}|A|\in \{\,A \in \mathrm{B}(\mathbf{A}) \mid c_{other} \in A\,\} & \\
	 & \forall c \in \mathrm{B}(\mathbf{C}), \mathrm{solution}(c) =x \Leftrightarrow c \in \exists A \in \mathrm{B}(\mathbf{A})                                   &
\end{alignat}

\subsubsection{Sukoro}\label{subsubsection:sukoro}

Possessed structure is $\mathbf{A}$. \textit{domain} and \textit{hidden} are
\begin{alignat}{4}
	 & \mathbf{P}   & \leftrightarrow \dcomma & \{\,null\,\}     \dcomma      & \rightarrow \dcomma & \{\,null\,\}      \\
	 & \mathbf{C}   & \leftrightarrow \dcomma & \{\,null,1,\dots 4\,\}\dcomma & \rightarrow \dcomma & \{\,undecided\,\} \\
	 & \mathbf{E_p} & \leftrightarrow \dcomma & \{\,null\,\}      \dcomma     & \rightarrow \dcomma & \{\,null\,\}      \\
	 & \mathbf{E_c} & \leftrightarrow \dcomma & \{\,null\,\}     \dcomma      & \rightarrow \dcomma & \{\,null\,\}      \\
	 & \mathbf{A}   & \leftrightarrow \dcomma & \{\,null\,\}     \dcomma      & \rightarrow \dcomma & \{\,null\,\}.
\end{alignat}

\textit{constraints} are
\begin{alignat}{2}
	 & |\mathrm{B}(\mathbf{A})|=1
	 &                                                                                                                                       \\
	 & \forall c \in \mathrm{B}(\mathbf{C}), \mathrm{solution}(c)\in \mathbb{N} \Leftrightarrow c \in \exists A \in \mathrm{B}(\mathbf{A}) & \\
	 & \land \mathrm{solution}(c)=|      \{\,x \in  \mathrm{connect}(c,\{\,H,V\,\})  \mid \mathrm{solution}(x)\in \mathbb{N}\,\}|          & \\
	 & \land \forall c_{other}\in \mathrm{connect}(c,\{\,H,V\,\}), \mathrm{solution}(c_{other})\neq \mathrm{solution}(c)                   &
\end{alignat}

\subsubsection{Norinori}\label{subsubsection:norinori}

Possessed structures are $\mathbf{A_1}$, $\mathbf{A_2}$ (both from \cref{equation:area}). \textit{domain} and \textit{hidden} are
\begin{alignat}{4}
	 & \mathbf{P}   & \leftrightarrow \dcomma & \{\,null\,\}     \dcomma  & \rightarrow \dcomma & \{\,null\,\}      \\
	 & \mathbf{C}   & \leftrightarrow \dcomma & \{\,null,x\,\}\dcomma     & \rightarrow \dcomma & \{\,undecided\,\} \\
	 & \mathbf{E_p} & \leftrightarrow \dcomma & \{\,null\,\}      \dcomma & \rightarrow \dcomma & \{\,null\,\}      \\
	 & \mathbf{E_c} & \leftrightarrow \dcomma & \{\,null\,\}     \dcomma  & \rightarrow \dcomma & \{\,null\,\}      \\
	 & \mathbf{A_1} & \leftrightarrow \dcomma & \{\,null\,\}     \dcomma  & \rightarrow \dcomma & \{\,null\,\}      \\
	 & \mathbf{A_2} & \leftrightarrow \dcomma & \{\,null\,\}     \dcomma  & \rightarrow \dcomma & \{\,null\,\}.
\end{alignat}

\textit{constraints} are
\begin{alignat}{2}
	 & \mathrm{no\_overlap}(\mathbf{A_1})                                                                                             & \\
	 & \mathrm{fill}(\mathbf{A_2})                                                                                                    & \\
	 & \forall c \in \mathrm{B}(\mathbf{C}), \mathrm{solution}(c) = x \Leftrightarrow c \in \exists A_1  \in \mathrm{B}(\mathbf{A_1}) & \\
	 & \forall A_1 \in \mathrm{B}(\mathbf{A_1}) , |A_1|=2
	 &                                                                                                                                  \\
	 & \forall A_2 \in \mathrm{B}(\mathbf{A_2}) , |\{\,c\in A_2 \mid \mathrm{solution}(c)=x\,\}            |=2                        &
\end{alignat}

\subsubsection{Hitori}\label{subsubsection:hitori}

Possessed structures are $\mathbf{A_h}$, $\mathbf{A_v}$, $\mathbf{A}$. \textit{domain} and \textit{hidden} are
\begin{alignat}{4}
	 & \mathbf{P}   & \leftrightarrow \dcomma & \{\,null\,\}     \dcomma  & \rightarrow \dcomma & \{\,null\,\}     \\
	 & \mathbf{C}   & \leftrightarrow \dcomma & \{\,1\dots n,x\,\}\dcomma & \rightarrow \dcomma & \{\,1\dots n\,\} \\
	 & \mathbf{E_p} & \leftrightarrow \dcomma & \{\,null\,\}      \dcomma & \rightarrow \dcomma & \{\,null\,\}     \\
	 & \mathbf{E_c} & \leftrightarrow \dcomma & \{\,null\,\}     \dcomma  & \rightarrow \dcomma & \{\,null\,\}     \\
	 & \mathbf{A_h} & \leftrightarrow \dcomma & \{\,null\,\}     \dcomma  & \rightarrow \dcomma & \{\,null\,\}.    \\
	 & \mathbf{A_v} & \leftrightarrow \dcomma & \{\,null\,\}     \dcomma  & \rightarrow \dcomma & \{\,null\,\}.    \\
	 & \mathbf{A}   & \leftrightarrow \dcomma & \{\,null\,\}     \dcomma  & \rightarrow \dcomma & \{\,null\,\}.
\end{alignat}

\textit{constraints} are
\begin{alignat}{2}
	                                                                                          & n=m                                                                                                                                   & \\
	                                                                                          & \mathrm{fill}(\mathbf{A_h})                                                                                                           & \\
	                                                                                          & \mathrm{fill}(\mathbf{A_v})                                                                                                           & \\
	                                                                                          & \forall A_h \in \mathrm{B}(\mathbf{A_h}), A'_h=\{\,c\in A_h\mid \mathrm{solution}(c)\neq x\,\} , \mathrm{all\_different}(A'_h)        & \\
	                                                                                          & \forall A_v \in \mathrm{B}(\mathbf{A_v}), A'_v=\{\,c\in A_v\mid \mathrm{solution}(c)\neq x\,\} , \mathrm{all\_different}(A'_v)        & \\
	                                                                                          & |\mathrm{B}(\mathbf{A})     |=1                                                                                                       & \\
	                                                                                          & \forall c \in \mathrm{B}(\mathbf{C}),\mathrm{solution}(c)=x \Leftrightarrow \{\,c\in A \mid A\in \mathrm{B}(\mathbf{A})\,\}=\emptyset & \\
	                                                                                          &
	\land \{\,y\in\mathrm{connect}(c,\{\,H,V\,\}) \mid \mathrm{solution}(y)=x \,\} =\emptyset &
\end{alignat}

\end{document}